\documentclass[10pt,twocolumn,letterpaper]{article}

\usepackage[pagenumbers]{cvpr}
\usepackage{times}
\usepackage{epsfig}
\usepackage{graphicx}
\usepackage{amsmath}
\usepackage{amssymb}

\usepackage{multirow}
\usepackage{array}
\usepackage{booktabs}
\usepackage{makecell}
\usepackage{mathtools}
\usepackage{afterpage}

\usepackage[accsupp]{axessibility}
\usepackage[pagebackref=true,breaklinks=true,letterpaper=true,colorlinks,bookmarks=false]{hyperref}

\usepackage[capitalize]{cleveref}
\crefname{section}{Sec.}{Secs.}
\Crefname{section}{Section}{Sections}
\Crefname{table}{Table}{Tables}
\crefname{table}{Tab.}{Tabs.}

\begin{document}

\title{Multi-View Depth Estimation by Fusing \\ Single-View Depth Probability with Multi-View Geometry}

\author{Gwangbin Bae \;\;\;\; Ignas Budvytis \;\;\;\; Roberto Cipolla\\
University of Cambridge\\
{\tt\small \{gb585,ib255,rc10001\}@cam.ac.uk}}
\maketitle

\begin{abstract}
Multi-view depth estimation methods typically require the computation of a multi-view cost-volume, which leads to huge memory consumption and slow inference. Furthermore, multi-view matching can fail for texture-less surfaces, reflective surfaces and moving objects. For such failure modes, single-view depth estimation methods are often more reliable. To this end, we propose MaGNet, a novel framework for fusing single-view depth probability with multi-view geometry, to improve the accuracy, robustness and efficiency of multi-view depth estimation. For each frame, MaGNet estimates a single-view depth probability distribution, parameterized as a pixel-wise Gaussian. The distribution estimated for the reference frame is then used to sample per-pixel depth candidates. Such probabilistic sampling enables the network to achieve higher accuracy while evaluating fewer depth candidates. We also propose depth consistency weighting for the multi-view matching score, to ensure that the multi-view depth is consistent with the single-view predictions. The proposed method achieves state-of-the-art performance on ScanNet \cite{ScanNet}, 7-Scenes \cite{7-scenes} and KITTI \cite{KITTI}. Qualitative evaluation demonstrates that our method is more robust against challenging artifacts such as texture-less/reflective surfaces and moving objects. Our code and model weights are available at \url{https://github.com/baegwangbin/MaGNet}.
\end{abstract}

\begin{figure*}[t]
\begin{center}
\includegraphics[width=1.0\linewidth]{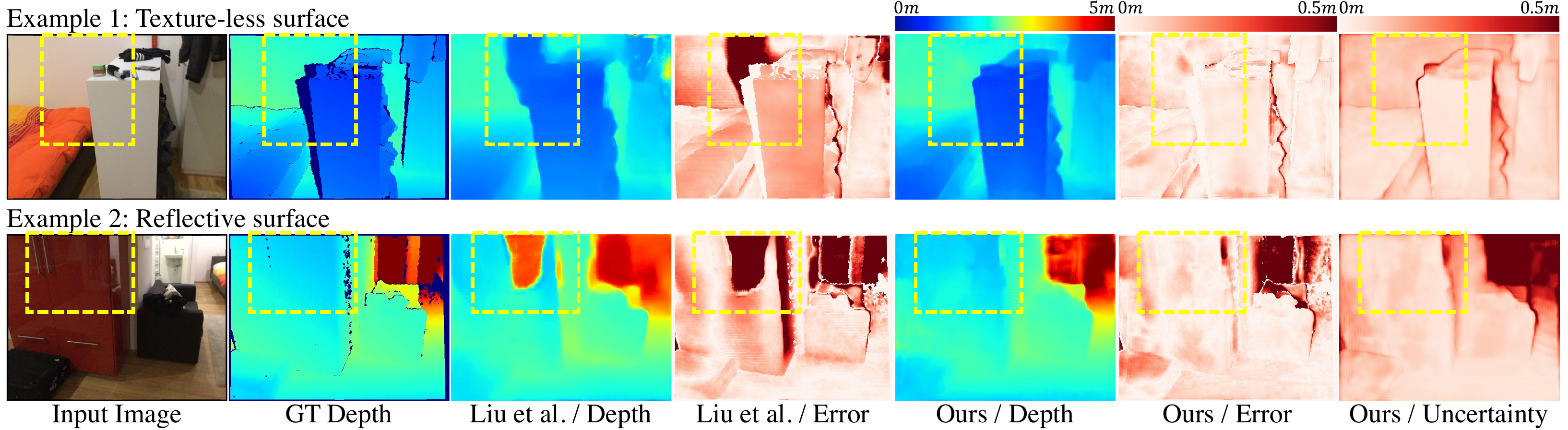}
\end{center}
\caption{This figure shows examples of images that are challenging for the existing multi-view depth estimation methods, such as \cite{videoP-2019-NeuralRGBD}. Multi-view matching can be unreliable if the scene contains texture-less or reflective surfaces, leading to inaccurate predictions (see yellow boxes). On the contrary, we use single-view depth probability to constrain the search space for depth candidates and to encode the depth consistency of each candidate in each view, resulting in more accurate and robust predictions.}
\label{fig:intro}
\end{figure*}

\section{Introduction}
\label{sec:intro}

Depth estimation is pivotal to 3D scene reconstruction and understanding. Owing to the advances in deep convolutional neural networks, many attempts have been made to estimate the pixel-wise metric depth from RGB images. Both single-view and multi-view methods have been proposed. The two families of solutions rely on different cues and therefore inherit different strengths and weaknesses.

Single-view methods \cite{mono-2014-eigen,mono-2015-eigen,mono-2015-liu,mono-2016-laina,mono-2018-DORN,mono-2019-VNL,mono-2019-BTS,mono-2020-adabins,mono-2021-dpt,mono-2021-transdepth} use \textit{monocular cues}, such as texture gradients and objects with known size. A deep feature extractor (e.g. \cite{VGGNET,RESNET}) is used to encode such cues into a dense feature map, from which a decoder regresses the per-pixel depth. With suitable supervision, single-view methods can learn the depth of weakly-textured or reflective surfaces. However, their accuracy is limited due to the inherent ambiguity of the problem.

Multi-view methods \cite{mvs-adaptive-2020-ATV,mvs-adaptive-2020-CascadeMVS,mvs-adaptive-2020-CVPMVS,videoP-2019-NeuralRGBD,video-2020-cnmnet,video-2021-long}, on the other hand, use \textit{geometric cues}. The key assumption adopted by these methods is that if the estimated depth for a particular pixel is correct, it will be projected to visually similar pixels in the other images. While such hard-coded multi-view geometry reduces the ambiguity and leads to better accuracy, there are several limitations: A large number of depth candidates should be evaluated in order for the correct depth to be found; The multi-view consistency assumption is violated in the presence of occlusion and object motion; Lastly, the multi-view matching becomes unreliable for texture-less or reflective surfaces.

We argue that both \textit{monocular} and \textit{geometric} cues should be exploited in order to complement the limitation of one another. The ambiguity in single-view depth can be reduced by performing multi-view matching. The efficiency of multi-view matching can be improved by sampling the depth candidates near the single-view depth. The failure cases of multi-view matching (e.g. on texture-less/reflective surfaces) can be prevented by enforcing the consistency with the single-view depth.

To this end, we introduce MaGNet (\textbf{M}onocular \textbf{a}nd \textbf{G}eometric \textbf{Net}work), a novel framework for fusing single-view depth probability with multi-view geometry. MaGNet uses a sequence of monocular images with known intrinsics and camera poses as input. The forward pass consists of the following steps: (1) The network estimates the single-view depth probability distribution of each image, parameterized as a pixel-wise Gaussian; (2) For each pixel in the reference image, a small number of depth candidates are sampled from the estimated depth probability distribution; (3) The sampled candidates are projected to the neighboring views and the matching scores are measured in terms of the dot product between the feature vectors; (4) The matching score computed for each neighboring view is multiplied by a binary depth consistency weight inferred from the single-view depth probability estimated from that viewpoint; (5) Lastly, the resulting \textit{thin} cost-volume is used to obtain a more accurate multi-view depth probability distribution. Steps (2-5) can be repeated to yield a more accurate result. The final output of our network is a map of per-pixel depth probability distribution, from which the expected value and the associated uncertainty can be inferred. 

Our contributions can be summarized as below.

\begin{itemize}
    \item \textbf{Probabilistic depth sampling.} Most multi-view depth estimation methods
    \cite{mvs-2018-DeepMVS,mvs-2018-MVDepthNet,mvs-2018-MVSNet,mvs-2019-DPSNET,mvs-2019-MVSCRF,mvs-2019-P-MVSNET,mvs-2019-R-MVSNet,videoP-2019-NeuralRGBD,video-2020-DeepV2D,video-2020-cnmnet,video-2021-long} use the same set of depth candidates (sampled between some hand-picked limits $d_\text{min}$ and $d_\text{max}$) for all pixels. Even the methods with coarse-to-fine depth search strategy \cite{mvs-adaptive-2020-ATV,mvs-adaptive-2020-CascadeMVS,mvs-adaptive-2020-CVPMVS} use uniformly sampled candidates to obtain the initial coarse depth-map. To achieve higher accuracy for lower computational cost, we propose probabilistic depth sampling, where per-pixel candidates are sampled from the single-view depth probability distribution. While \cite{mvs-2019-DPSNET,video-2020-cnmnet, mvs-2018-MVDepthNet,videoP-2019-NeuralRGBD,video-2021-long} evaluate 64 uniformly sampled candidates, we only sample 5 candidates (i.e. 92\% thinner cost-volume).

    \item \textbf{Depth consistency weighting for multi-view matching.} We use the single-view depth probability estimated in each view to encode the depth consistency of the candidates. By multiplying the multi-view matching score with a binary depth consistency weight, we improve the robustness and accuracy.

    \item \textbf{Iterative refinement.} The result of the probabilistic depth sampling and consistency-weighted multi-view matching is a \textit{thin} cost-volume, which is used to update the initial depth probability distribution. However, if the initial single-view depth probability distribution is inaccurate, none of the sampled depth candidates will be near the true depth. To handle such failure mode, we introduce iterative refinement where the updated distribution is fed back to the probabilistic depth sampling module. Ablation study shows that such iterative refinement leads to higher accuracy.
\end{itemize}

Experimental results show that MaGNet achieves state-of-the-art performance on ScanNet \cite{ScanNet}, 7-Scenes \cite{7-scenes} and KITTI \cite{KITTI}. Qualitative evaluation shows that the network is more robust against challenging artifacts such as reflective and texture-less surfaces (see Fig. \ref{fig:intro}).

\section{Related Work}

\noindent
\textbf{Monocular depth estimation.} Despite the inherently ill-posed nature of the problem, monocular depth estimation has been studied extensively in literature. While early learning-based approaches \cite{mono-2006-saxena,mono-2008-saxena} relied on hand-crafted image features, recent approaches \cite{mono-2014-eigen,mono-2015-eigen,mono-2015-liu,mono-2016-laina,mono-2018-DORN,mono-2019-BTS,mono-2019-VNL,mono-2020-adabins,mono-2021-dpt,mono-2021-transdepth} predict depth from CNN features. Notable contributions have been made by recasting depth estimation as an ordinal regression problem \cite{mono-2018-DORN}, introducing virtual normal loss to enforce geometric constraints \cite{mono-2019-VNL}, or using vision transformers \cite{other-ViT1,other-ViT2} to encode the global context \cite{mono-2020-adabins,mono-2021-transdepth,mono-2021-dpt}. 

\noindent
\textbf{Multi-view depth estimation.} When given a sequence of monocular images with known intrinsics and camera poses, multi-view stereo (MVS) \cite{mvs-2015-furukawa,mvs-2016-COLMAP} can be used to estimate the per-pixel depth. Learning-based MVS methods \cite{mvs-2018-DeepMVS,mvs-2018-MVDepthNet,mvs-2018-MVSNet,mvs-2019-DPSNET,mvs-2019-MVSCRF,mvs-2019-P-MVSNET,mvs-2019-PointMVSNET,mvs-2019-R-MVSNet,mvs-2020-Fast-MVSNET,mvs-adaptive-2020-ATV,mvs-adaptive-2020-CascadeMVS,mvs-adaptive-2020-CVPMVS} sample per-pixel depth candidates, project them to the neighboring views and measure their matching scores (between CNN features) to infer the per-pixel depth. State-of-the-art methods are generally evaluated on DTU \cite{DTU} and Tanks and Temples \cite{Tanks_and_Temples}. Both datasets were captured in a controlled setup, where the camera and the depth sensor were kept static or gimbal-stabilized. On the contrary, datasets like ScanNet \cite{ScanNet}, 7-Scenes \cite{7-scenes} and KITTI \cite{KITTI} are captured by sensors attached to a hand-held device or a moving vehicle. The images often contain motion blur, texture-less/reflective surfaces and moving objects, all of which make the multi-view matching challenging. Methods like \cite{videoP-2019-NeuralRGBD, video-2020-cnmnet, video-2020-deltas, mvs-2020-NormalAssisted, video-2021-long} focus on such datasets. They use surface normal as additional supervisory signal \cite{video-2020-cnmnet,mvs-2020-NormalAssisted} or enforce the spatio-temporal consistency between multiple frames \cite{videoP-2019-NeuralRGBD,video-2021-long}. These methods will be the main competitors of our approach.

\noindent
\textbf{Coarse-to-fine depth sampling.} Most multi-view depth estimation methods use the same set of depth candidates for all pixels. To obtain high accuracy, the candidates should be sampled densely (e.g. \cite{mvs-2019-MVSCRF,mvs-2019-P-MVSNET} use 256 candidates), leading to huge memory consumption and slow inference. To overcome such limitation, recent MVS methods \cite{mvs-adaptive-2020-ATV,mvs-adaptive-2020-CascadeMVS,mvs-adaptive-2020-CVPMVS} use coarse-to-fine strategy to construct a multi-scale cost-volume. Firstly, the depth candidates are sampled uniformly to obtain a coarse cost-volume. Then, the candidates for higher resolution are sampled near the coarse depth-map. While \cite{mvs-adaptive-2020-CascadeMVS} halved the search space in each iteration, \cite{mvs-adaptive-2020-ATV} introduced uncertainty-based adaptive sampling, where the per-pixel variance is inferred from the coarse cost-volume to define the search space.

\noindent
\textbf{Probabilistic depth estimation.} In order to deploy CNN-based depth estimation methods in safety-critical applications, the networks should not only be accurate but also be able to quantify the uncertainty in prediction. Two major types of uncertainty are aleatoric and epistemic \cite{monoP-2017-Kendall}. Aleatoric uncertainty (i.e. uncertainty in data) in depth is commonly learned by estimating the probability distribution over possible depths. Both discrete and continuous solutions have been proposed. Discrete solutions \cite{monoP-2017-cao,monoP-2019-Yang,videoP-2019-NeuralRGBD} formulate depth estimation as classification over discretized depths and hence suffer from the quantization error. Continuous solutions \cite{monoP-2017-Kendall} represent depth probability as a parameterized distribution (e.g. Gaussian) and train the network by maximizing the likelihood of the ground truth. We take a step further and demonstrate how the single-view depth probability distributions estimated from different views can be used to obtain a more accurate multi-view distribution.


\begin{figure*}[t]
\begin{center}
\includegraphics[width=1.0\linewidth]{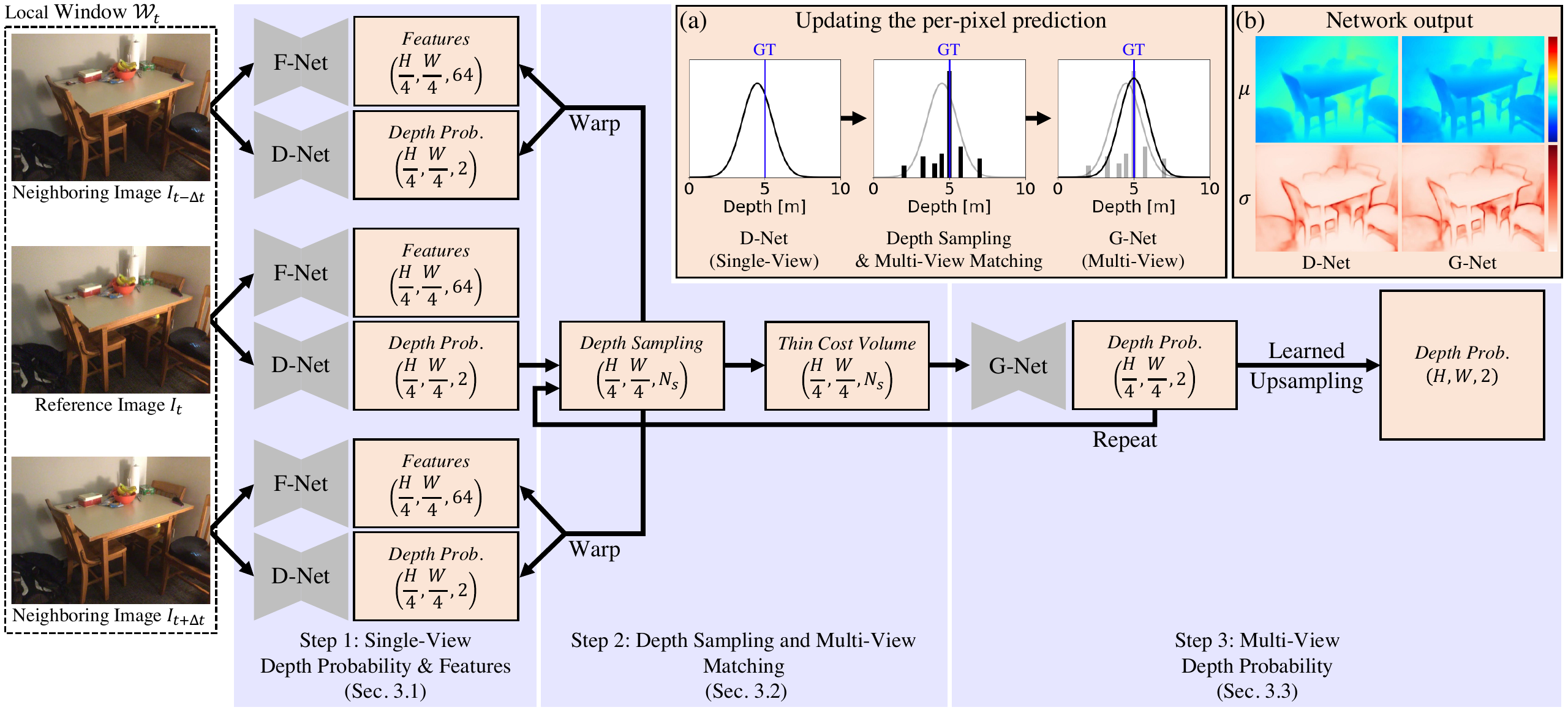}
\end{center}
\caption{This figure illustrates the proposed pipeline. For each image, D-Net estimates the single-view depth probability and F-Net extracts features. D-Net output for the reference frame is used to sample per-pixel depth candidates, which are evaluated via the consistency-weighted multi-view matching. From the obtained thin cost-volume, G-Net updates the mean and variance of the initial depth probability distribution, which can be fed back to the depth sampling module to yield a more accurate prediction. \textbf{(a)} This figure illustrates how the per-pixel prediction is updated. The curves and histograms represent the estimated depth probability distributions and the matching scores of the sampled candidates, respectively. \textbf{(b)} This figure shows the update in the dense prediction (depth $\mu$ and uncertainty $\sigma$).}
\label{fig:method}
\end{figure*}

\section{Method}
\label{sec:method}

Our goal is to estimate a depth-map for the reference frame $I_t$ at time $t$. The input to the network is a local window of images $\mathcal{W}_t = \{ I_{t-2\Delta t}, I_{t-\Delta t}, I_{t}, I_{t+\Delta t}, I_{t+2\Delta t} \}$ with known intrinsics and camera poses. The proposed pipeline, illustrated in Fig. \ref{fig:method}, consists of three steps: For each image, the network estimates the single-view depth probability distribution and extracts features (Sec. \ref{sec:method1}); The estimated single-view depth probability is fused with multi-view geometry via probabilistic depth sampling and consistency weighting (Sec. \ref{sec:method2}); Lastly, the resulting thin cost-volume is used to estimate the multi-view depth probability distribution (Sec. \ref{sec:method3}).

\subsection{Single-View Depth Probability and Features}
\label{sec:method1}

\noindent
\textbf{Single-view depth probability.} For each image in $\mathcal{W}_t$ which has the resolution of $H \times W$, D-Net estimates a map of single-view depth probability distribution in reduced resolution of $H/4 \times W/4$. The distribution for each pixel $(u,v)$ in the input image $I_t$ is parameterized as a Gaussian,
\begin{align}
    \label{eqn:single-view-dpv}
    p_{u,v}(d|I_t)
    =
    \frac{1}{\sigma_{u,v}(I_t) \sqrt{2\pi}}
    e^{
    -\frac{1}{2}
    \left(
    \frac{d - \mu_{u,v}(I_t)}{\sigma_{u,v}(I_t)}
    \right)^2},
\end{align}
\noindent
where $\mu$ and $\sigma^2$ are the mean and the variance. Any existing depth estimation network can be used as D-Net. We use a lightweight convolutional encoder-decoder with EfficientNet B5 \cite{EfficientNet} backbone. We use linear activation for $\mu$, and the modified ELU function \cite{other-ELU}, $f(x)=\text{ELU}(x)+1$ for $\sigma^2$ to ensure positive variance and smooth gradient. D-Net is pre-trained and the weights are fixed when training the other components of the pipeline. The training loss is the negative log-likelihood (NLL) of the ground truth depth,
\begin{align}
\label{eqn:dnet-loss}
L_{u,v}(d^\text{gt}_{u,v}|I_t)
=
\frac{1}{2} \log \sigma^2_{u,v}(I_t)
+ 
\frac{\left( 
d^\text{gt}_{u,v} - \mu_{u,v}(I_t)
\right)^2
}{2 \sigma^2_{u,v}(I_t)}.
\end{align}

%

Eq. \ref{eqn:dnet-loss} is an L2 loss with learned attenuation. The network learns to estimate high $\sigma^2$ when it is challenging to reduce the error $(d^\text{gt} - \mu)^2$. This generally happens near object boundaries and for distant points \cite{monoP-2017-Kendall}. On the contrary, when the estimated $\sigma^2$ is low, the correct depth is likely to be near the estimated $\mu$. We will explain in Sec. \ref{sec:method2} how such information can be exploited to improve the efficiency and accuracy of multi-view matching.

\noindent
\textbf{Single-view features.} For each image, F-Net extracts a feature map of resolution $H/4 \times W/4$. We use the architecture of \cite{other-PSMNET} as in \cite{videoP-2019-NeuralRGBD}. Following \cite{other-DeepPruner}, the matching score between two pixels is computed in terms of the dot product between the feature vectors. For pixel $(u,v)$ with depth candidates $\{d_k\}^{N_s}_{k=1}$, the matching score can be written as

\begin{equation}
\begin{aligned}
\label{eqn:feature-metric}
s_{u,v,k}(I_{t}) 
&= \sum_{i \neq t}
\langle 
\mathbf{f}_{u,v}(I_t), \mathbf{f}_{u_{ik},v_{ik}}(I_i) 
\rangle,
\end{aligned}
\end{equation}

\noindent
where $\langle \cdot , \cdot \rangle$ represents the dot product and $(u_{ik},v_{ik})$ is the projection of the 3D coordinates defined by $(u,v,d_k)$ on the $i$-th image. By applying softmax, the cost-volume can be transformed into a depth probability volume, $p_{u,v,k} = \text{softmax}_k s_{u,v,k}$, from which the expected per-pixel depth can be inferred as $\hat{d}_{u,v} = \sum_k p_{u,v,k} \cdot d_k$. F-Net is also pre-trained by using uniformly sampled depth candidates $\{d_k\}$ and minimizing the L1 loss between $\hat{d}_{u,v}$ and $d_{u,v}^\text{gt}$. 

\subsection{Fusing Single-View Depth Probability with Multi-View Geometry}
\label{sec:method2}

In this section, we explain how single-view depth probability can be fused with multi-view geometry. The components described in this section have no learnable parameters.

\noindent
\textbf{Probabilistic depth sampling.} The single-view depth probability distribution estimated for the reference frame is used to sample per-pixel depth candidates. Firstly, we define the search space $[\mu_{u,v} - \beta \sigma_{u,v},\mu_{u,v} + \beta \sigma_{u,v}]$ for each pixel where $\beta$ is a hyper-parameter. Then, we split the interval into $N_s$ bins such that each bin shares the same amount of probability mass. This ensures that more candidates are sampled near $\mu_{u,v}$ (i.e. the most likely depth value). The mid-point of each bin is then selected as a depth candidate. The $k$-th depth candidate, $d_{u,v,k}$, is thus defined as

\begin{equation}
\begin{aligned}
\label{eqn:sampling}
d_{u,v,k} 
&= \mu_{u,v} + b_{k} \sigma_{u,v}, \\
\text{where}\;\;\;\;\;
b_k &= \frac{1}{2} \bigg[ \Phi^{-1} \left( \frac{k-1}{N_s} P^* 
+ \frac{1-P^*}{2} \right) 
\\ &\;\;\;\;\;\;\;\;\; + \Phi^{-1} \left( \frac{k}{N_s} P^* + \frac{1-P^*}{2} \right) \bigg]. \\
\end{aligned}
\end{equation}

\begin{figure}[t]
\begin{center}
\includegraphics[width=1.0\linewidth]{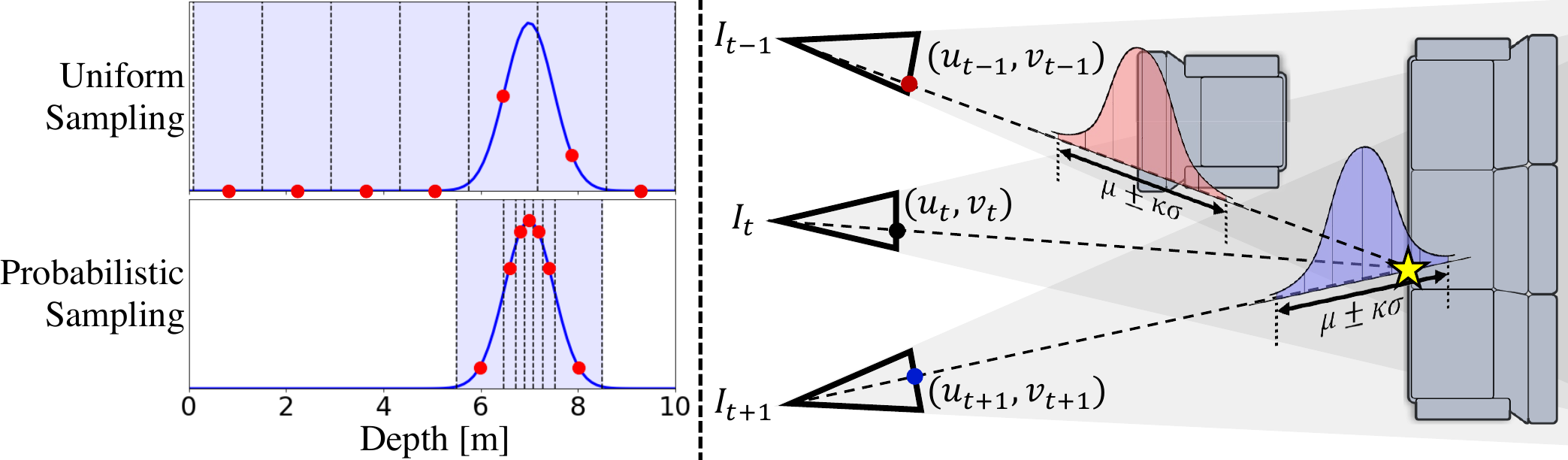}
\end{center}
\caption{\textbf{(left)} Comparison between uniform sampling and the proposed probabilistic sampling. The blue curve represents the single-view depth probability distribution, and the red dots represent the sampled candidates. \textbf{(right)} Illustration of depth consistency weighting. For pixel $(u_t, v_t)$ in the reference frame $I_t$, a depth candidate defines a 3D point (marked with $\star$). This point is projected to the neighboring views and the depth probability in each view is evaluated. For $I_{t-1}$, $\star$ is not within $\mu \pm \kappa \sigma$ due to occlusion. The consistency weight becomes 0 in such case.}
\label{fig:sampling_and_consistency}
\end{figure}

In Eq. \ref{eqn:sampling}, $\Phi^{-1}(\cdot)$ is the probit function and $P^*=\text{erf}(\beta / \sqrt{2})$ is the probability mass covered by the interval $[\mu_{u,v} \pm \beta \sigma_{u,v}]$ (see supplementary material for detailed derivation). Note that the values of $\{ b_k \}$ only depend on $N_s$ and $\beta$ (i.e. they are not calculated per-pixel). Fig. \ref{fig:sampling_and_consistency}-(left) compares the proposed sampling against the uniform sampling. As we only sample within the $\beta$-sigma confidence interval, we can achieve higher accuracy while evaluating fewer candidates. For the pixels with high uncertainty, the spacing between the candidates increases so that a wider range of candidates can be evaluated. 

\noindent
\textbf{Depth consistency weighting.} If a depth candidate is correct, this means that the corresponding 3D point is on the surface of some scene element (e.g. objects). If this 3D point is visible in some neighboring view, the corresponding single-view depth probability (estimated from that view) should be high. Assuming that this is true (i.e. assuming our D-Net is accurate), the logically equivalent contrapositive is "if the single-view depth probability of a depth candidate estimated from a neighboring view is low, it means either that the depth candidate is wrong or that it is not visible in that view (e.g. due to occlusion)". The multi-view matching score should not be computed for such case (see Fig. \ref{fig:sampling_and_consistency}-(right)). To this end, we introduce a binary weighting for the multi-view matching score,
\begin{equation}
\begin{aligned}
\label{eqn:weighted-feature-metric}
s_{u,v,k}(I_t)
&= \sum_{i \neq t}
w^{\text{dc}}_{u_{ik},v_{ik},d_{ik}}
\langle 
\mathbf{f}_{u,v}(I_t), \mathbf{f}_{u_{ik},v_{ik}}(I_i) 
\rangle \\
w^{\text{dc}}_{u_{ik},v_{ik},d_{ik}} &= \delta 
\left( p_{u_{ik},v_{ik}}(d_{ik}|I_i) > p_{\text{thres}}
\right).
\end{aligned}
\end{equation}

In Eq. \ref{eqn:weighted-feature-metric}, $w^{\text{dc}}_{u_{ik},v_{ik},d_{ik}}$ is 1 if the single-view depth probability $p_{u_{ik},v_{ik}}(d_{ik}|I_i)$ evaluated from the $i$-th image is above certain threshold $p_\text{thres}$ and is 0 otherwise. We call this depth consistency weighting. Setting the right $p_\text{thres}$ is important. If it is too high, it will zero out too many depth candidates, one of which can be the correct one. We set $p_{\text{thres}}=\exp(-\kappa^2/2)/\sigma_{u_{ik}, v_{ik}} \sqrt{2\pi}$ so that the weight becomes 1 if $d_{ik}$ is within the $\kappa$-sigma confidence interval. This means that $p_{\text{thres}}$ is adaptive both per-pixel and per-view. If D-Net is uncertain about the depth (i.e. high $\sigma$), $p_{\text{thres}}$ becomes low, allowing more depth candidates to be considered. 

Depth consistency weighting discards the candidates with low single-view depth probability. Such weighting is useful especially when the multi-view matching is ambiguous or unreliable. For example, if the pixel is within a texture-less surface, a wide range of depth candidates will lead to similar matching scores. If the scene contains reflective surfaces, the matching score will be computed between the \textit{reflections}, resulting in over-estimated depth. In both cases, MaGNet can make robust prediction by favoring the depth candidates with high single-view depth probability. 

\subsection{Estimating Multi-View Depth Probability Distribution}
\label{sec:method3}

\noindent
\textbf{Updating single-view depth probability distribution.} The result of the probabilistic depth sampling and consistency-weighted multi-view matching is a \textit{thin} cost-volume of size $H/4 \times W/4 \times N_s$, where $N_s$ is the number of depth candidates. Using this as an input, G-Net estimates the \textit{multi-view} depth probability distribution by updating the mean and variance of the initial single-view distribution. Each element of the cost-volume, $s_{u,v,k}$, is a matching score computed at pixel $(u,v)$ for the $k$-th depth candidate, $\mu_{u,v} + b_k \sigma_{u,v}$ (see Eq. \ref{eqn:sampling}). Since the values of $\mu_{u,v}$ and $\sigma_{u,v}$ are not encoded in the input, it is difficult to directly regress the updated mean and variance.

Instead, our G-Net estimates the normalized residual $\Delta \mu_{u,v} / \sigma_{u,v}$. For example, if the matching score is high for the $k'$-th depth candidate, the network should predict $b_{k'}$, so that the updated mean becomes $\mu^{\text{new}}_{u,v} = \mu_{u,v} + b_{k'} \sigma_{u,v}$. Similarly, G-Net also estimates $\sigma^{\text{new}}_{u,v} / \sigma_{u,v}$ to update the variance. This gives us the updated, multi-view depth probability distribution $\mathcal{N}(\mu^\text{new}_{u,v}, \sigma^\text{new}_{u,v})$ for each pixel. Note that the output of G-Net can be fed back to the sampling module and the process can be repeated to refine the output.

\noindent
\textbf{Learned upsampling.} The output of G-Net is a map of multi-view depth probability distribution of resolution $H/4 \times W/4$. To recover the full resolution, we use the learned upsampling layer introduced in \cite{other-RAFT}. The input to the layer is the feature-map of D-Net (see supplementary material for the network architecture). A light-weight CNN estimates $H/4 \times W/4 \times (4\times 4\times 9)$ mask and the full resolution depth at each pixel is computed as the weighted sum of the $3\times 3$ grid of its coarse resolution neighbors. 

\noindent
\textbf{Iterative refinement and network training.} The multi-view matching process (i.e. probabilistic depth sampling $\rightarrow$ consistency-weighted matching $\rightarrow$ update by G-Net) is repeated for $N_\text{iter}$ times, producing $N_\text{iter}$ predictions. For each prediction, the NLL loss (Eq. \ref{eqn:dnet-loss}) is computed, and their sum is used to train G-Net and the upsampling layer. Following \cite{other-RAFT}, the $i$-th prediction is weighted by $\gamma^{N_\text{iter} - i}$, where $0 < \gamma < 1$, to put bigger emphasis on the final output. 

Iterative refinement is beneficial in two regards. Firstly, if one of the candidates achieves a high matching score, the mean will be shifted towards that candidate and the variance will be reduced, so that in the next iteration, the network can perform a \textit{finer} depth search near that candidate to find a better candidate with a higher matching score. Iterative update can also prevent the failure mode where the D-Net prediction is inaccurate. For example, if the true depth is not within the initial search space $[\mu_{u,v} - \beta \sigma_{u,v},\mu_{u,v} + \beta \sigma_{u,v}]$, none of the sampled candidates will achieve a high matching score. In such case, G-Net will learn to increase the variance to attenuate the loss (Eq. \ref{eqn:dnet-loss}), and the network can perform a \textit{wider} depth search in the next iteration.


\begin{table*}[t]
\setlength{\tabcolsep}{2.6pt}
\begin{center}
\begin{tabular}{l|c|ccccc|ccccc}
\toprule
\multirow{2}{4em}{Method} & \multirow{2}{*}{Cap} 
& \multicolumn{5}{c|}{Train on ScanNet $\rightarrow$ Test on ScanNet} 
& \multicolumn{5}{c}{Train on ScanNet $\rightarrow$ Test on 7-Scenes}\\
\cline{3-12}
 & 
& abs rel & abs diff & rmse & rmse$_{\log}$ & {\scriptsize $\delta\!\!<\!\!1.25$}
& abs rel & abs diff & rmse & rmse$_{\log}$ & {\scriptsize $\delta\!\!<\!\!1.25$} \\
\midrule
MVDepthNet \cite{mvs-2018-MVDepthNet} 
& \multirow{8}{*}{10m}
& 0.1116 & 0.2087 & 0.3143 & 0.1500 & 88.04 & 0.1905 & 0.3304 & 0.4260 & 0.2221 & 71.93 \\
DPSNet \cite{mvs-2019-DPSNET} &
& 0.0986 & 0.1998 & 0.2840 & 0.1348 & 88.80 & 0.1675 & 0.2970 & 0.3905 & 0.2061 & 76.03 \\
NAS \cite{mvs-2020-NormalAssisted} &
& 0.0941 & 0.1928 & 0.2703 & 0.1269 & 90.09 & 0.1631 & 0.2885 & 0.3791 & 0.1997 & 77.12 \\
CNM-Net \cite{video-2020-cnmnet} &
& 0.1102 & 0.2129 & 0.3032 & 0.1482 & 86.88 & 0.1602 & 0.2751 & 0.3602 & 0.2030 & 76.81 \\
DELTAS \cite{video-2020-deltas} &
& 0.0915 & 0.1710 & 0.2390 & 0.1226 & 91.47 & 0.1548 & 0.2671 & 0.3541 & 0.1860 & 79.66 \\
UCS-Net \cite{mvs-adaptive-2020-ATV} &
& 0.0845 & 0.1605 & 0.2335 & 0.1145 & 92.22
& 0.2113 & 0.3668 & 0.4683 & 0.2369 & 69.31 \\
Long et al. \cite{video-2021-long} &
& 0.0812 & 0.1505 & 0.2199 & 0.1104 & \textbf{93.13} & 0.1465 & 0.2528 & 0.3382 & 0.1967 & 80.36 \\
\cline{1-1}\cline{3-12}
Ours (D-Net) & 
& 0.1186 & 0.2070 & 0.2708 & 0.1461 & 85.46 
& 0.1339 & 0.2209 & \textbf{0.2932} & 0.1677 & 83.08 \\
Ours (full) & 
& \textbf{0.0810} & \textbf{0.1466} & \textbf{0.2098} & \textbf{0.1101} & 92.98 
& \textbf{0.1257} & \textbf{0.2133} & 0.2957 & \textbf{0.1639} & \textbf{85.52} \\
\hline
\hline
NeuralRGBD \cite{videoP-2019-NeuralRGBD} & \multirow{4}{*}{5m}
& 0.1013 & 0.1657 & 0.2500 & 0.1315 & 91.60 & 0.2334 & 0.4060 & 0.5358 & 0.2516 & 68.03 \\
Long et al. \cite{video-2021-long} &
& 0.0805 & 0.1438 & 0.2029 & \textbf{0.1083} & \textbf{93.33} & 0.1465 & 0.2528 & 0.3382 & 0.1967 & 80.36 \\
\cline{1-1}\cline{3-12}
Ours (D-Net) & 
& 0.1177 & 0.1991 & 0.2526 & 0.1439 & 85.70 
& 0.1339 & 0.2209 & \textbf{0.2932} & 0.1677 & 83.08 \\
Ours (full) & 
& \textbf{0.0804} & \textbf{0.1409} & \textbf{0.1960} & 0.1084 & 93.13 
& \textbf{0.1257} & \textbf{0.2133} & 0.2957 & \textbf{0.1639} & \textbf{85.52} \\
\bottomrule
\end{tabular}
\end{center}
\caption{Quantitative evaluation on ScanNet \cite{ScanNet} and 7-Scenes \cite{7-scenes}. We follow the evaluation protocol of \cite{video-2021-long}. While the accuracy of MaGNet on ScanNet is similar to that of \cite{video-2021-long}, we show superior generalization ability, where we outperform other methods on all metrics.}
\label{table:benchmark}
\end{table*}


\section{Experimental Setup}

\noindent
\textbf{Dataset and evaluation protocol.} We train MaGNet on ScanNet \cite{ScanNet}. ScanNet contains 2.7M views from 1613 scans. We use the official data split to train and test the model. To evaluate the generalization ability, we perform a cross-dataset evaluation on the test split of the 7-Scenes dataset \cite{7-scenes} without fine-tuning. We also train and test our method on KITTI \cite{KITTI}, both using the Eigen split \cite{mono-2014-eigen} and the official split. For all evaluations, depth accuracy is measured using the metrics defined in \cite{mono-2014-eigen}. 

\noindent
\textbf{Implementation details.} MaGNet is implemented using PyTorch \cite{PyTorch}. We first train D-Net and F-Net (separately), and fix their weights when training the remaining components. We use AdamW optimizer \cite{other-AdamW} and schedule the learning rate using 1cycle policy \cite{other-1cycle-lr} with $lr_\text{max} = 3.5 \times 10^{-4}$. The batch size is 16/4/8 for D-, F- and G-Net (plus the upsampling layer), respectively. The number of epochs is 5/2/2 for ScanNet \cite{ScanNet} and 10/5/5 for KITTI \cite{KITTI}. For indoor datasets, we use a local window of five images where $\Delta t$ is set to ten frames. For KITTI \cite{KITTI}, we use three images and $\Delta t$ is set to two frames. The hyper-parameters are $\{ \beta, \kappa, \gamma  \} = \{ 3.0, 5.0, 0.8 \} $ in all experiments. $N_s$ and $N_\text{iter}$ are 5 and 3, unless specified otherwise. 


\begin{table}[t]
\small
\setlength{\tabcolsep}{2.4pt}
\begin{center}
\begin{tabular}{l|c|ccccc}
\toprule
Method & Multi & abs rel & sq rel & rmse & rmse$_{\log}$ 
& {\scriptsize $\delta\!\!<\!\!1.25$} \\
\midrule
MonoDepth2 \cite{self-MS-2019-monodepth2} & $\times$
& 0.106 & 0.806 & 4.630 & 0.193 & 87.6 \\
FeatDepth \cite{self-MS-2020-feature} & $\times$
& 0.099 & 0.697 & 4.427 & 0.184 & 88.9 \\
BTS \cite{mono-2019-BTS} & $\times$
& 0.059 & 0.245 & 2.756 & 0.096 & 95.6 \\
AdaBins \cite{mono-2020-adabins} & $\times$
& 0.058 & 0.190 & 2.360 & 0.088 & 96.4 \\
SC-GAN \cite{video-2019-SCGAN} & $\checkmark$
& 0.063 & 0.178 & \textbf{2.129} & 0.097 & 96.1 \\
\hline
Ours (D-Net) & $\times$
& 0.061 & 0.209 & 2.422 & 0.092 & 96.0 \\
Ours (full) & $\checkmark$
& \textbf{0.054} & \textbf{0.162} & 2.158 & \textbf{0.083} & \textbf{97.1} \\
\hline
\hline
NeuralRGBD \cite{videoP-2019-NeuralRGBD} & $\checkmark$
& 0.100 & 0.473 & 2.829 & 0.128 & 93.2 \\
\hline
Ours (D-Net) & $\times$
& 0.063 & 0.254 & 2.471 & 0.102 & 95.8 \\
Ours (full) & $\checkmark$
& \textbf{0.050} & \textbf{0.167} & \textbf{1.971} & \textbf{0.085} & \textbf{97.7} \\
\bottomrule
\end{tabular}
\end{center}
\caption{Quantitative evaluation on KITTI \cite{KITTI}. The second column shows whether the method operates on a multi-view setup. We used the Eigen split \cite{mono-2014-eigen} except for the comparison against \cite{videoP-2019-NeuralRGBD}, where we used the official split. Our method shows state-of-the-art performance.}
\label{table:bm-kitti}
\end{table}

\begin{figure*}[t]
\begin{center}
\includegraphics[width=0.97\linewidth]{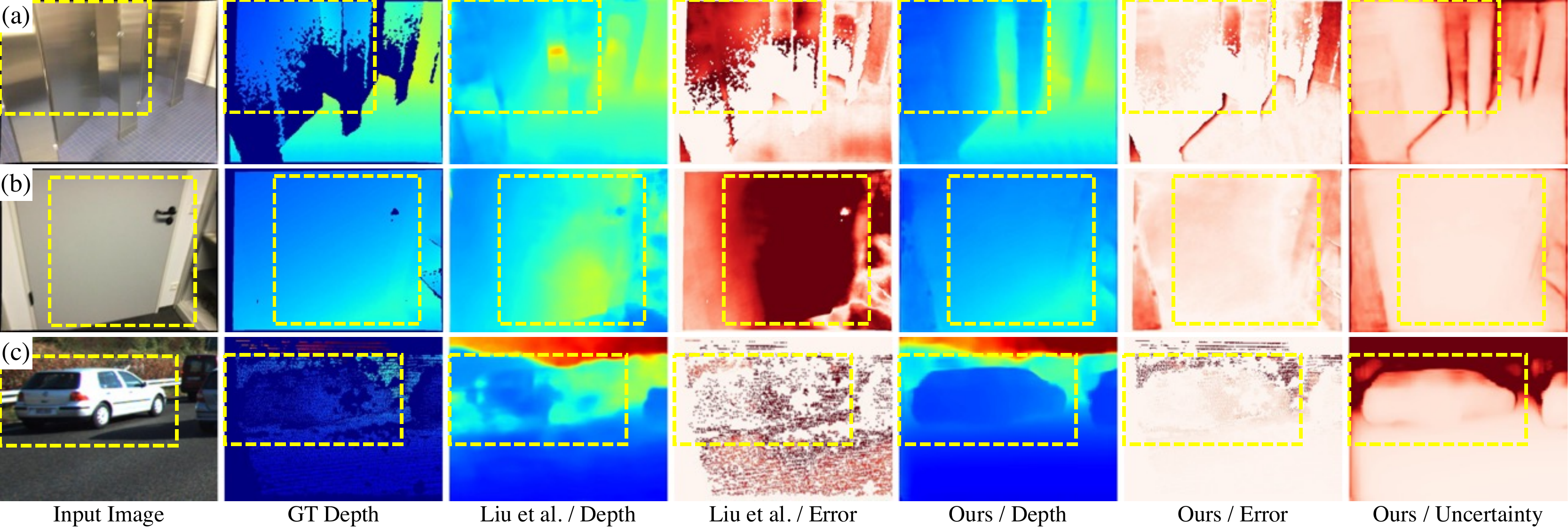}
\end{center}
\caption{Qualitative comparison against \cite{videoP-2019-NeuralRGBD}. With the proposed fusion of single-view depth probability, MaGNet can make accurate prediction for (a) reflective surfaces, (b) weakly-textured surfaces and (c) moving objects. The estimated uncertainty also correlates well with the prediction error. See supplementary material for more examples.}
\label{fig:benchmark}
\end{figure*}

\section{Experiments}
\label{sec:exp}

\subsection{Comparison with the State-of-the-Art}
\label{sec:exp1}

\noindent
\textbf{ScanNet and 7-Scenes.} Tab. \ref{table:benchmark} shows that MaGNet achieves state-of-the-art performance on both ScanNet \cite{ScanNet} and 7-Scenes \cite{7-scenes}. While the accuracy on ScanNet is similar to \cite{video-2021-long}, our method shows superior generalization ability. State-of-the-art methods \cite{videoP-2019-NeuralRGBD,video-2020-cnmnet,video-2021-long} operate on a huge cost-volume covering the entire depth range (e.g. 0-10m). In such case, the networks may learn the characteristics specific to the dataset (e.g. the depths of the pixels can in general be small/large for certain dataset), leading to over-fitting. On the contrary, MaGNet operates on a thin cost-volume, where the per-pixel entries cover a small depth range of $\mu_{u,v} \pm \beta \sigma_{u,v}$. The low dimensionality of the input makes the network less prone to over-fitting. Qualitative comparison against \cite{videoP-2019-NeuralRGBD} (Fig. \ref{fig:benchmark}) shows that MaGNet is more robust against challenging artifacts such as reflective/texture-less surfaces and moving objects. Note that while \cite{videoP-2019-NeuralRGBD} evaluates 64 depth candidates per pixel, we only evaluate 15 (5 candidates $\times$ 3 iterations). 

\noindent
\textbf{KITTI.} Tab. \ref{table:bm-kitti} shows that MaGNet outperforms the state-of-the-art methods on KITTI \cite{KITTI}. KITTI is a challenging dataset for multi-view depth estimation methods for two reasons: (1) The images often contain moving objects, for which multi-view consistency is violated; (2) The camera generally moves in a forward direction, resulting in a small baseline (i.e. less accurate multi-view matching). However, as MaGNet uses single-view depth to restrict the depth search space and to enforce the depth consistency, it is more robust against such artifacts, as can be seen in Fig. \ref{fig:benchmark}.

\subsection{Ablation Study}
\label{sec:exp2}

In this section, we perform ablation studies to confirm the effectiveness of the proposed probabilistic depth sampling, depth consistency weighting, and iterative refinement. Note that the accuracy is reported on a smaller test set of ScanNet \cite{ScanNet} provided by \cite{mono-2018-DORN}.

\begin{figure}[t]
\begin{center}
\includegraphics[width=1.0\linewidth]{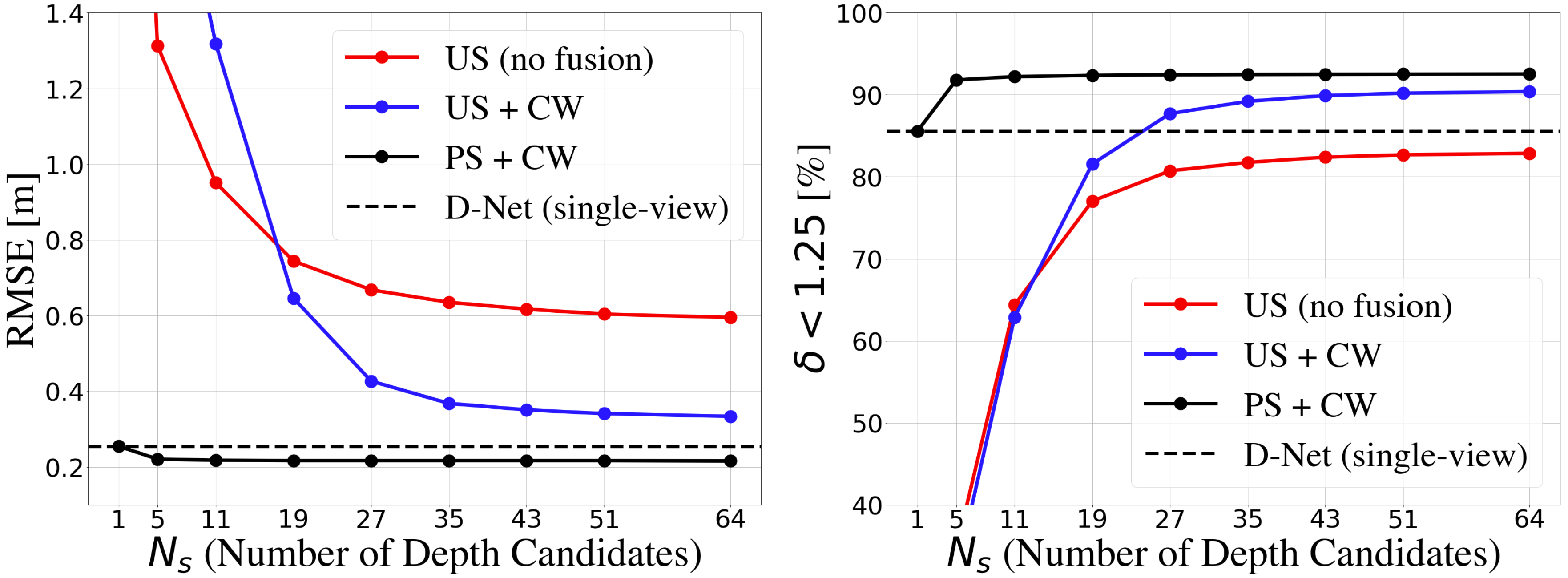}
\end{center}
\caption{Effectiveness of the proposed fusion of single-view depth probability. US, PS and CW mean Uniform Sampling, Probabilistic Sampling and Consistency Weighting. Our full model (PS+CW) achieves higher accuracy while evaluating fewer candidates. Dashed line in each plot shows the accuracy of D-Net (single-view). Our full model becomes equivalent to D-Net when $N_s=1$. Without the proposed sampling and weighting, the RMSE cannot be lower than that of D-Net, even for large $N_s$.}
\label{fig:ablation-qt}
\end{figure}

\begin{figure}[t]
\begin{center}
\includegraphics[width=1.0\linewidth]{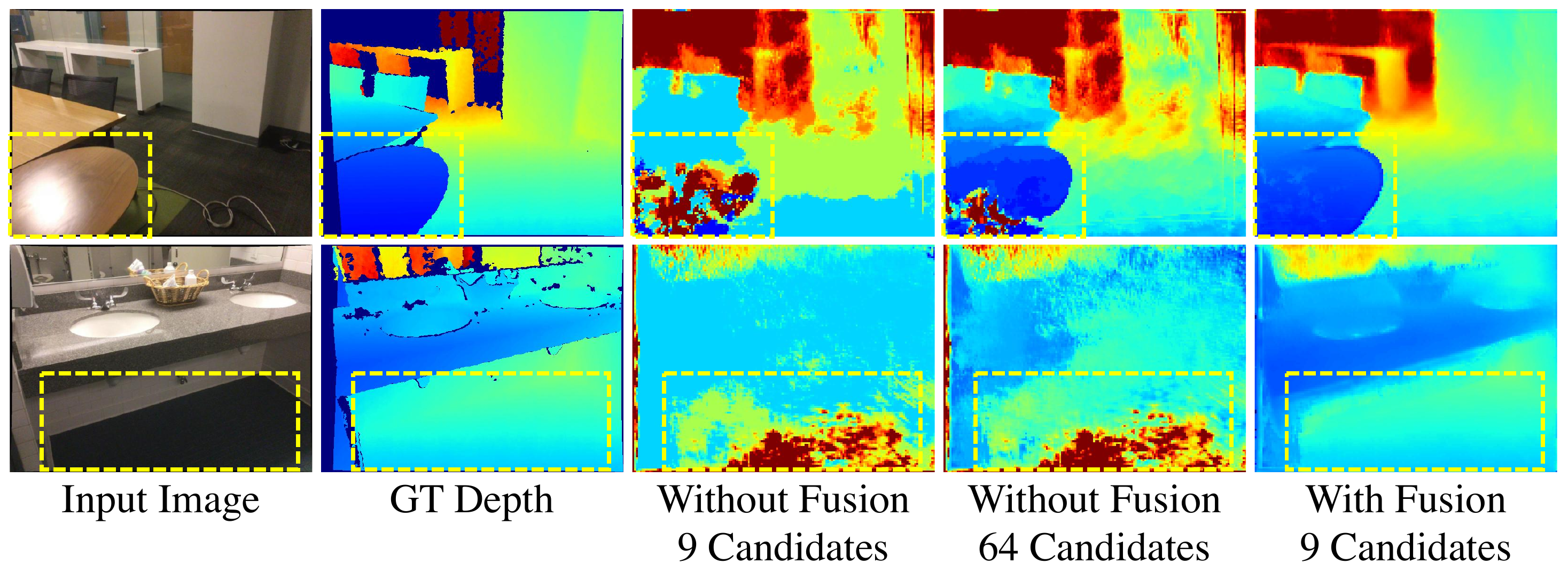}
\end{center}
\caption{The proposed fusion of single-view depth probability makes the multi-view matching more robust against challenging artifacts, such as reflective surfaces (top) and texture-less surfaces (bottom).}
\label{fig:ablation-ql}
\end{figure}

\noindent
\textbf{Effectiveness of the proposed fusion of single-view depth probability.} We compare the accuracy of multi-view matching with and without the proposed probabilistic sampling and consistency weighting. To ensure a fair comparison, the accuracy is evaluated directly from the cost-volume (by applying softmax and solving $\hat{d}_{u,v} \!=\! \sum_k p_{u,v,k} \cdot d_k$). Fig. \ref{fig:ablation-qt} shows that both components lead to significant improvement in the accuracy. By fusing single-view depth probability, it is possible to achieve higher accuracy while evaluating fewer candidates. Note that the consistency weighting alone can improve the accuracy (for $N_s \geq 19$). This suggests that the proposed weighting can be applied to the existing multi-view depth estimation methods that operate on uniformly sampled candidates. Qualitative comparison in Fig. \ref{fig:ablation-ql} shows that the proposed fusion makes the multi-view matching more robust against challenging artifacts, such as reflective and texture-less surfaces.

\begin{table}[t]
\small
\setlength{\tabcolsep}{4pt}
\begin{center}
\begin{tabular}{c|c|ccccc}
\toprule
$N_\text{iter}$ & $N_\text{s}$ & abs rel & sq rel & rmse & rmse$_{\log}$ & $\delta\!\!<\!\!1.25$ \\
\midrule
\multirow{4}{*}{1} & 5 & 0.097 & 0.035 & 0.217 & 0.121 & 90.75 \\
 & 7 & 0.096 & 0.035 & 0.217 & 0.121	& 90.81 \\
 & 9 & 0.096 & 0.035 & 0.217 & 0.121	& 90.81 \\
 & 11 & \textbf{0.095} & \textbf{0.034} & \textbf{0.216} & \textbf{0.120} & \textbf{90.94} \\
\hline
1 & \multirow{4}{*}{5} & 0.097 & 0.035 & 0.217 & 0.121 & 90.75 \\
2 & & 0.090 & 0.032 & 0.209 & 0.115 & 92.15 \\
3 & & \textbf{0.087} & 0.031 & 0.207 & \textbf{0.113} & 92.61 \\
4 & & \textbf{0.087} & \textbf{0.030} & \textbf{0.206} & \textbf{0.113} 
& \textbf{92.73} \\
\bottomrule
\end{tabular}
\end{center}
\caption{Accuracy for different values of $N_\text{iter}$ (number of iterations) and $N_s$ (number of depth candidates).
It is better to repeat the multi-view matching multiple times with small $N_s$ than to perform a single matching with large $N_s$. The accuracy converges for $N_\text{iter} \geq 3$.}
\label{table:ablation-sampling}
\end{table}

\noindent
\textbf{Iterative refinement.} We also report the accuracy of the full pipeline for different values of $N_s$ (number of depth candidates) and $N_\text{iter}$ (number of iterations) in Tab. \ref{table:ablation-sampling}. Since the depth candidates are concentrated near the estimated mean, the spacing between the candidates are small even for low $N_s$. As a result, increasing $N_s$ does not lead to meaningful improvement in the accuracy. On the contrary, repeating the multi-view matching process leads to significant improvement. If the initial multi-view matching is successful (i.e. one of the candidates achieves a high matching score), the network can perform a \textit{finer} search in the next iteration. If it is unsuccessful, the variance increases and the network can perform a \textit{wider} search in the next iteration. In summary, it is better to repeat the process multiple times with small $N_s$ than to perform a single iteration with large $N_s$. The increase in the computational cost is small as the multi-view matching is performed in low resolution ($H/4 \times W/4$) and for a small number of samples. For $N_s=5$, each iteration takes 11.23ms on a single 2080Ti GPU. The accuracy converges for $N_\text{iter} \geq 3$.

\noindent
\textbf{Comparison against cascade cost volume-based MVS.} Cascade cost volume-based MVS methods \cite{mvs-adaptive-2020-ATV,mvs-adaptive-2020-CascadeMVS,mvs-adaptive-2020-CVPMVS} use coarse-to-fine depth sampling. UCS-Net \cite{mvs-adaptive-2020-ATV} uses uncertainty-based sampling and is thus similar to our method. The difference is two-fold. Firstly, UCS-Net requires 64 uniformly sampled candidates to estimate the initial coarse depth-map, whereas MaGNet only samples 5 candidates from the single-view depth probability distribution. Secondly, UCS-Net performs the next multi-view matching in a higher resolution, while MaGNet stays in the coarse resolution, evaluating only 5$\times$2 additional samples (i.e. more memory efficient). Tab. \ref{table:ucs-net} compares the two methods. We also train UCS-Net by replacing the 64 uniformly sampled initial candidates with 8 candidates sampled from our D-Net prediction ("UCS-Net + PS"). With the help of the proposed probabilistic sampling, UCS-Net can achieve similar accuracy, while being significantly faster and lighter (e.g. when trained on four 2080Ti GPUs, the training speed increases from 17fps to 31fps and memory consumption reduces from 10.3GB/gpu to 5.5GB/gpu). Our full model (with consistency weighting and iterative refinement) performs better than both variants of \cite{mvs-adaptive-2020-ATV}.

\noindent
\textbf{Limitations.} MaGNet uses single-view depth probability distributions to (1) sample depth candidates and (2) infer their depth consistency. However, single-view depth, due to its inherent ambiguity, can be inaccurate. This is why we designed both components to be uncertainty-aware. We also proposed iterative multi-view matching, where G-Net updates the variance so that the depth sampling can be \textit{finer} or \textit{wider} in the next iteration. As a result, MaGNet can handle mild inaccuracy in single-view predictions. Nevertheless, the proposed pipeline can suffer in \textit{cross-domain evaluations}. For example, if MaGNet is trained on ScanNet \cite{ScanNet} (indoor) and tested on KITTI \cite{KITTI} (outdoor), the sampling range (i.e. $\mu \pm \beta \sigma$) will not include the true depth, even after multiple updates by G-Net. This is mainly because it is difficult for a single-view network to infer the metric scale of the scene. A possible solution can be to train D-Net with a scale-invariant loss \cite{mono-2014-eigen}, so that it can estimate the \textit{relative} depth. Then, the scaling factor for each image can be obtained by minimizing the reprojection error. This will be addressed in our future work.

\begin{table}[t]
\small
\setlength{\tabcolsep}{1.0pt}
\begin{center}
\begin{tabular}{l|c|ccccc}
\toprule
Method & $N_s$ & abs rel & sq rel & rmse & rmse$_{\log}$ & {\scriptsize $\delta\!\!<\!\!1.25$} \\
\midrule
UCS-Net \cite{mvs-adaptive-2020-ATV}
& (64, 32, 8)
& 0.091 & 0.156 & 0.229 & 0.120 & 91.49 \\
UCS-Net + PS
& (8, 8, 8) 
& 0.092 & 0.157 & 0.214 & 0.118 & 90.74 \\
\hline
Ours
& (5$\times$3, 0, 0)
& \textbf{0.082} & \textbf{0.143} & \textbf{0.202} & \textbf{0.110}
& \textbf{92.78} \\
\bottomrule
\end{tabular}
\end{center}
\caption{Comparison against cascade cost volume-based approach \cite{mvs-adaptive-2020-ATV}. 
$N_s$ represents the number of samples in each resolution ($H/4 \times W/4$, $H/2 \times W/2$, and $H \times W$). Unlike \cite{mvs-adaptive-2020-ATV}, MaGNet stays in the coarse resolution ($H/4 \times W/4$). Replacing the initial uniform sampling in coarse resolution with the proposed probabilistic sampling (PS) results in similar accuracy, while significantly reducing the training time and memory consumption.}
\label{table:ucs-net}
\end{table}

\section{Conclusion}

In this paper, we proposed a novel framework for fusing single-view depth probability with multi-view geometry to improve the accuracy, efficiency and robustness of multi-view depth estimation. Specifically, we introduced probabilistic depth sampling where per-pixel depth candidates are sampled from the single-view depth probability distribution, and depth consistency weighting for the multi-view matching score to ensure that the multi-view depth is consistent with the single-view predictions. We also proposed iterative multi-view matching, where a small number of candidates are sampled from the current depth probability distribution to update its mean and variance. The proposed method shows state-of-the-art performance on ScanNet \cite{ScanNet}, 7-Scenes \cite{7-scenes} and KITTI \cite{KITTI}. Ablation study illustrates that the proposed fusion of single-view depth probability improves the accuracy, efficiency and robustness of multi-view depth estimation.

\noindent 
\textbf{Acknowledgement.} This research was sponsored by Toshiba Europe's Cambridge Research Laboratory.

\appendix

\section{Mathematical Formulation}
\label{sec:maths}

In this section, we provide the mathematical formulation for the proposed probabilistic depth sampling and consistency-weighted multi-view matching.

\subsection{Probabilistic Depth Sampling}

Suppose that, for pixel $(u,v)$ in the reference image, D-Net estimates the single-view depth probability distribution (parameterized as a Gaussian) of mean $\mu_{u,v}$ and variance $\sigma^2_{u,v}$. We first define the search space $[\mu_{u,v} - \beta \sigma_{u,v}, \mu_{u,v} + \beta \sigma_{u,v}]$, where $\beta$ is a hyper-parameter. Given the probability density function $p_{u,v}(x)$, the probability mass $P^*_{u,v}$ covered by the search space is
\begin{equation}
\begin{aligned}
\label{eqn:eq1}
P^*_{u,v} 
&= \int^{\mu_{u,v}+\beta\sigma_{u,v}}_{\mu_{u,v}-\beta\sigma_{u,v}} p_{u,v}(x)dx \\
&= F_{u,v} (\mu_{u,v} + \beta\sigma_{u,v}) - F_{u,v} (\mu_{u,v} - \beta\sigma_{u,v}) \\
&= \text{erf} \left( \frac{\beta}{\sqrt{2}} \right),
\end{aligned}
\end{equation}

\noindent
where $F_{u,v}(\cdot)$ and $\text{erf}(\cdot)$ are the cumulative distribution function and the error function. $P^*_{u,v}$ depends only on $\beta$ and we can thus drop the subscripts. We then split the search space into $N_s$ bins of equal probability mass, $P^*/N_s$, and select their mid-points as depth candidates. The $k$-th depth candidate, $d_{u,v,k}$, is thus defined as
\begin{equation}
\begin{aligned}
\label{eqn:depth-candidate}
d_{u,v,k} 
&= \frac{1}{2} 
\bigg[ F^{-1}_{u,v} 
\left( \frac{k-1}{N_s} P^* + \frac{1-P^*}{2} \right)
\\ &\;\;\;\;\;\;\;\; + 
F^{-1}_{u,v} 
\left( \frac{k}{N_s} P^* + \frac{1-P^*}{2} \right) \bigg], \\
\end{aligned}
\end{equation}

\noindent
where $F^{-1}_{u,v}(\cdot)$ is the Gaussian quantile function (i.e. the inverse of $F_{u,v}(\cdot)$). Eq. \ref{eqn:depth-candidate} can be simplified into Eq. \ref{eqn:sampling} by using the relation $F^{-1}_{u,v}(p) = \mu_{u,v} + \sigma_{u,v} \Phi^{-1}(p)$, where $\Phi^{-1}(p)$ is the quantile function of the standard normal distribution (i.e. the probit function).

\subsection{Consistency-Weighted Multi-View Matching}

In this section, we explain how the consistency-weighted multi-view matching score of each depth candidate can be calculated.

\noindent
\textbf{Step 1. Pixel coordinates of $I_t$ $\rightarrow$ World coordinates.} A 3D point $\mathbf{X}^c_t$ in the camera-centered coordinates of the reference image $I_t$ at time $t$ is projected to the pixel coordinates $\mathbf{w}_t$ via perspective projection. Given the camera calibration matrix $\mathbf{K}$, this can be written as
\begin{equation}
\begin{aligned}
\label{eqn:w=KX}
\tilde{\mathbf{w}}_t &= \mathbf{K} \mathbf{X}^c_t \\
\begin{bmatrix} su \\ sv \\ s \end{bmatrix}
&=
\begin{bmatrix} \alpha_u&0&u_0 \\ 0&\alpha_v&v_0 \\ 0&0&1 \end{bmatrix}
\begin{bmatrix} X^c_t \\ Y^c_t \\ Z^c_t \end{bmatrix},
\end{aligned}
\end{equation}

\noindent
where $\tilde{\mathbf{w}}_t$ is the homogeneous representation of the pixel coordinates $\mathbf{w}_t = (u,v)$. Suppose that, for pixel $(u,v)$ in the reference image $I_t$, we have sampled $N_s$ depth candidates $\{d_{k}\}^{N_s}_{k=1}$. Each depth candidate, together with the pixel coordinates, defines $\mathbf{X}^c_t$ as
\begin{equation}
\label{eqn:ray-eq}
\mathbf{X}^c_t = 
\begin{bmatrix} X^c_t \\ Y^c_t \\ Z^c_t \end{bmatrix} = 
\begin{bmatrix} \frac{u - u_0}{\alpha_u} \\ \frac{v - v_0}{\alpha_v} \\ 1 \end{bmatrix} \times d_k.
\end{equation}

The world coordinates of $\mathbf{X}^c_t$ is then given as $\tilde{\mathbf{X}}^w_t = \mathbf{P}^{-1}_{t} \tilde{\mathbf{X}}^c_t$, where $\mathbf{P}_{t}$ is the rigid body transformation matrix (i.e. camera pose) of $I_t$, and $\tilde{\mathbf{X}}$ is the homogeneous representation of $\mathbf{X}$.

\noindent
\textbf{Step 2. World coordinates $\rightarrow$ Pixel coordinates of $I_i$.} $\mathbf{X}^w_t$ is then projected to a neighboring image, $I_i$. The projection can be written as
\begin{equation}
\begin{aligned}
\label{eqn:to-nghbr}
\tilde{\mathbf{X}}^c_{i} &= \begin{bmatrix} X_{ik} \\ Y_{ik} \\ d_{ik} \\ 1 \end{bmatrix} = \mathbf{P}_{i} \tilde{\mathbf{X}}^w_t \\
\text{and} \;\;\;
\tilde{\mathbf{w}}_i &= \begin{bmatrix} su_{ik} \\ sv_{ik} \\ s \end{bmatrix} = \mathbf{K} 
\mathbf{X}^c_i.
\end{aligned}
\end{equation}

In Eq. \ref{eqn:to-nghbr}, $\mathbf{X}^c_i=(X_{ik}, Y_{ik}, d_{ik})$ is the camera-centered coordinates of image $I_i$ and $\mathbf{w}_i=(u_{ik},v_{ik})$ is the corresponding pixel coordinates. It is the projection of the 3D coordinates - defined by $(u,v)$ and $d_k$ in $I_t$ - on $I_i$. This defines $u_{ik}$, $v_{ik}$ and $d_{ik}$ that appear in Eq. \ref{eqn:feature-metric} and \ref{eqn:weighted-feature-metric}.

\noindent
\textbf{Step 3. Calculating the matching score.} The matching score for the $k$-th depth candidate is computed in terms of the dot product between the feature vectors - $\mathbf{f}_{u,v}(I_t)$ and $\mathbf{f}_{u_{ik},v_{ik}}(I_i)$. Since $u_{ik}$ and $v_{ik}$ are continuous values, $\mathbf{f}_{u_{ik},v_{ik}}(I_i)$ is obtained by bilinearly interpolating the 4-pixel neighbors - $(\lceil u_{ik} \rceil, \lceil v_{ik} \rceil)$, $(\lfloor u_{ik} \rfloor, \lceil v_{ik} \rceil)$, $(\lceil u_{ik} \rceil, \lfloor v_{ik} \rfloor)$ and $(\lfloor u_{ik} \rfloor, \lfloor v_{ik} \rfloor)$. 

\noindent
\textbf{Step 4. Consistency-weighting.} Similarly, $\mu_{u_{ik},v_{ik}}(I_i)$ and $\sigma_{u_{ik},v_{ik}}(I_i)$ can be bilinearly interpolated to evaluate the single-view depth probability of $d_{ik}$ at $(u_{ik}, v_{ik})$, which gives the depth consistency weight (Eq. \ref{eqn:weighted-feature-metric}).


\section{Additional Quantitative Results}
\label{sec:qt}

\begin{table}[t]
\setlength{\tabcolsep}{2.0pt}
\begin{center}
\begin{tabular}{l|c|c|c}
\toprule
\multirow{2}{4em}{Method} 
& \multicolumn{3}{c}{NLL (lower the better)} \\
\cline{2-4}
 & ScanNet & 7-Scenes & KITTI \\
\midrule
D-Net (single-view) & 2.235 & 2.794 & 2.644 \\
Full pipeline (multi-view) & \textbf{0.145} & \textbf{1.561} & \textbf{1.805} \\
\bottomrule
\end{tabular}
\end{center}
\caption{Negative log-likelihood of the ground truth depth evaluated for D-Net (single-view) and the full pipeline.}
\label{table:nll}
\end{table}

\noindent
\textbf{Negative log-likelihood.} Depth metrics (e.g. RMSE) only reflect the accuracy of the predicted mean $\mu$. In order to evaluate the accuracy of the distribution $\mathcal{N}(\mu,\sigma^2)$, we report the average negative log-likelihood (NLL) of the ground truth depth. Tab. \ref{table:nll} shows how NLL is reduced via multi-view matching.


\section{Additional Qualitative Results}
\label{sec:ql}

\begin{figure*}[t]
\begin{center}
\includegraphics[width=1.0\linewidth]{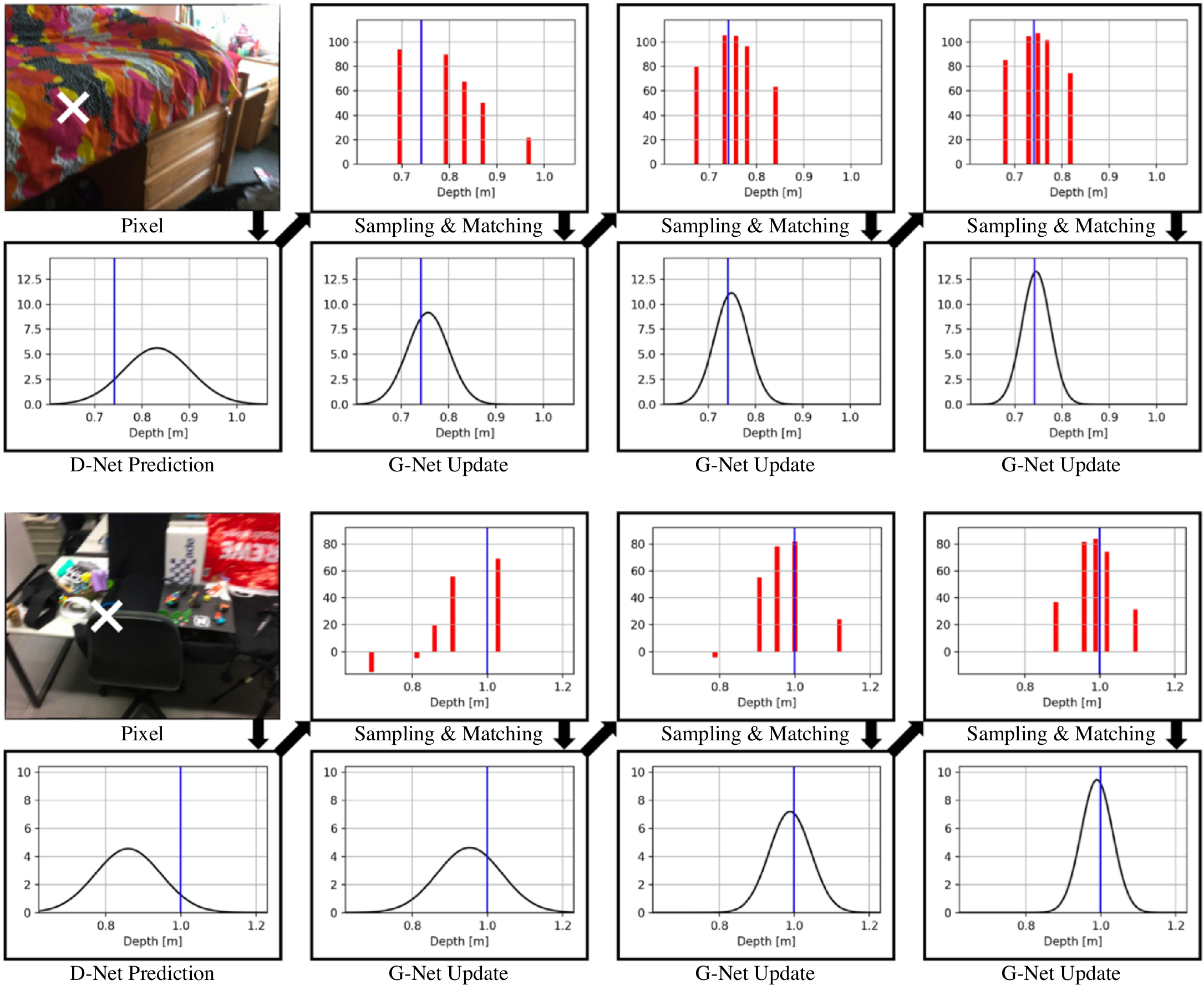}
\end{center}
\caption{Iterative update in the pixel-wise depth probability distribution. The black curves and blue vertical lines are the estimated depth probability distributions and the ground truth depth, respectively. Red bar plots show the consistency-weighted multi-view matching scores measured at the sampled depth candidates. Throughout the iterative refinement, the distribution becomes more accurate and confident (i.e. $\sigma$ is reduced).}
\label{fig:pixel}
\end{figure*}

\noindent
\textbf{Update in the pixel-wise prediction.} Fig. \ref{fig:pixel} shows how the pixel-wise prediction is updated throughout the pipeline. The initial D-Net prediction has high uncertainty, mainly due to the inherent ambiguity of single-view depth. This leads to large search space. After performing the multi-view matching for the sampled candidates, G-Net shifts the mean towards the candidate with high matching score. The variance is reduced so that the spacing between the candidates is smaller in the next iteration (i.e. better depth resolution).

\begin{figure*}[t]
\begin{center}
\includegraphics[width=1.0\linewidth]{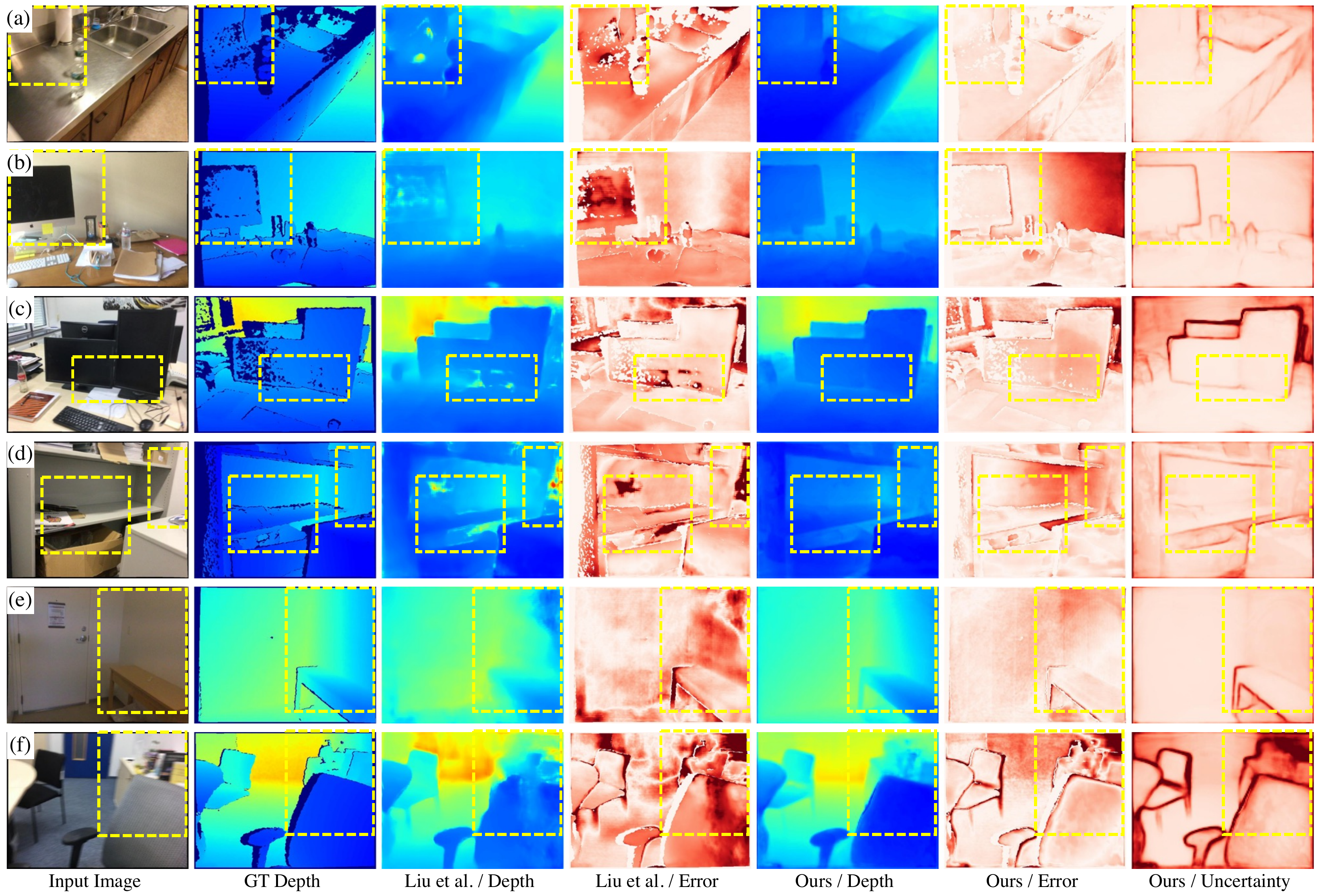}
\end{center}
\caption{Qualitative comparison against \cite{videoP-2019-NeuralRGBD} on ScanNet \cite{ScanNet}. With the proposed fusion of single-view depth probability (probabilistic sampling and consistency weighting), MaGNet can make accurate prediction for reflective surfaces (a-c) and texture-less surfaces (d-e). If a depth candidate is occluded in a particular neighboring frame, the corresponding single-view depth probability estimated from that view will be low. In such case, MaGNet ignores that view by setting the depth-consistency weight to zero. This improves the prediction near occlusion boundaries (f).}
\label{fig:bm-indoor}
\end{figure*}

\begin{figure*}[t]
\begin{center}
\includegraphics[width=1.0\linewidth]{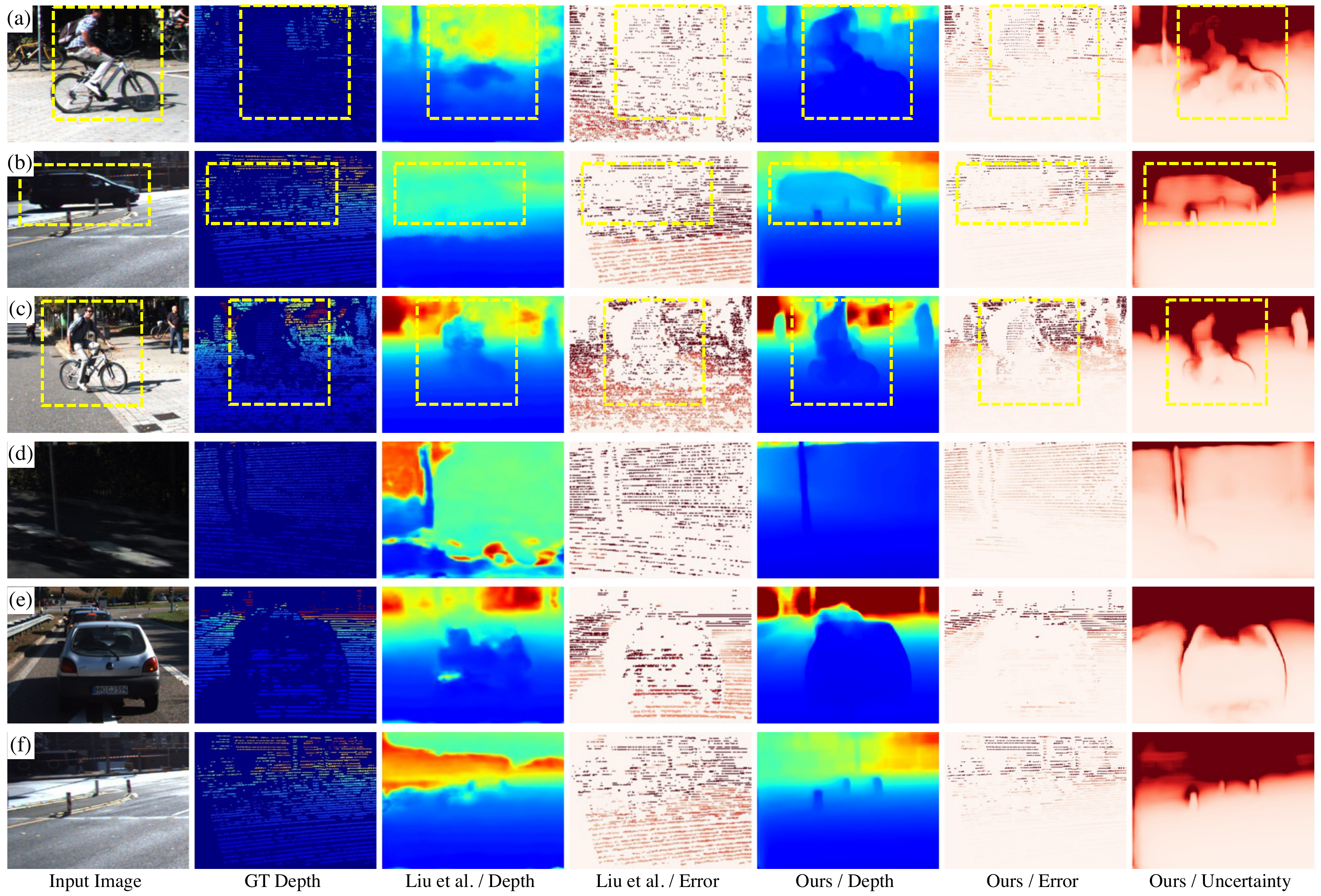}
\end{center}
\caption{Qualitative comparison against \cite{videoP-2019-NeuralRGBD} on KITTI \cite{KITTI}. Examples (a-c) show dynamic objects, for which multi-view consistency assumption is violated. Examples (d-f) show images where the camera was static within the local window. In such case, all depth candidates lead to same matching score (i.e. ambiguous matching). In both scenarios, MaGNet can make accurate predictions. This is because the proposed fusion enforces the final output to be consistent with the single-view depth probability distributions.}
\label{fig:bm-outdoor}
\end{figure*}

\noindent
\textbf{Qualitative comparison against the state-of-the-art.} Fig. \ref{fig:bm-indoor} and Fig. \ref{fig:bm-outdoor} provide additional qualitative comparison against \cite{videoP-2019-NeuralRGBD} (extension of Fig. \ref{fig:benchmark}). Our network shows superior performance especially on texture-less/reflective surfaces and moving objects.

\section{Network Architecture}
\label{sec:architecture}

\begin{table*}[t]
\setlength{\tabcolsep}{2.6pt}
\begin{center}
\begin{tabular}{c|c|c|c}
\hline
\textbf{Input} & \textbf{Layer} & \textbf{Output} & \textbf{Output Dimension} \\
\hline
\textit{image} & - & - & $H \times W \times 3$ \\
\hline
\multicolumn{4}{c}{\textbf{Encoder}} \\
\hline
\multirow{4}{*}{\textit{image}}
& \multirow{4}{*}{EfficientNet B5 \cite{EfficientNet}}
& \textit{$FEAT_4$} & $H/4 \times W/4 \times 40$ \\
& & \textit{$FEAT_8$} & $H/8 \times W/8 \times 64$ \\
& & \textit{$FEAT_{16}$} & $H/16 \times W/16 \times 176$ \\
& & \textit{$FEAT_{32}$} & $H/32 \times W/32 \times 2048$ \\
\hline
\multicolumn{4}{c}{\textbf{Decoder}} \\
\hline
\textit{$FEAT_{32}$} & Conv2D(ks=1, $C_\text{out}$=2048, padding=0) & \textit{x\_d0} & $H/32 \times W/32 \times 2048$ \\
\hline
upsample(\textit{x\_d0}) + \textit{$FEAT_{16}$} & 
$\left( 
\begin{matrix} 
\text{Conv2D(ks=3, $C_\text{out}$=1024, padding=1)}, \\
\text{BatchNorm2D()}, \\
\text{LeakyReLU()}
\end{matrix}
\right)
\times 2$ 
& \textit{x\_d1} & $H/16 \times W/16 \times 1024$ \\
\hline
upsample(\textit{x\_d1}) + \textit{$FEAT_{8}$} & 
$\left( 
\begin{matrix} 
\text{Conv2D(ks=3, $C_\text{out}$=512, padding=1)}, \\
\text{BatchNorm2D()}, \\
\text{LeakyReLU()}
\end{matrix}
\right)
\times 2$ 
& \textit{x\_d2} & $H/8 \times W/8 \times 512$ \\
\hline
upsample(\textit{x\_d2}) + \textit{$FEAT_{4}$} & 
$\left( 
\begin{matrix} 
\text{Conv2D(ks=3, $C_\text{out}$=256, padding=1)}, \\
\text{BatchNorm2D()}, \\
\text{LeakyReLU()}
\end{matrix}
\right)
\times 2$ 
& \textit{x\_d3} & $H/4 \times W/4 \times 256$ \\
\hline
\textit{x\_d3} & 
\makecell{
Conv2D(ks=3, $C_\text{out}$=128, padding=1), ReLU(), \\
Conv2D(ks=1, $C_\text{out}$=128, padding=0), ReLU(), \\
Conv2D(ks=1, $C_\text{out}$=2, padding=0) \\
}
& \textit{out} & $H/4 \times W/4 \times 2$ \\
\hline
\end{tabular}
\end{center}
\caption{D-Net architecture. In each convolutional layer, "ks" means the kernel size and $C_\text{out}$ is the number of output channels. $FEAT_N$ represents the feature-map of resolution $H/N \times W/N$. $X + Y$ means that the two tensors are concatenated, and $\text{upsample}(\cdot)$ is bilinear upsampling. Note that a different activation function is applied to each channel of the final output, as explained in Sec. \ref{sec:method1}.}
\label{table:dnet}
\end{table*}

\begin{table*}[t]
\setlength{\tabcolsep}{2.6pt}
\begin{center}
\begin{tabular}{c|c|c|c}
\hline
\textbf{Input} & \textbf{Layer} & \textbf{Output} & \textbf{Output Dimension} \\
\hline
\textit{cost-volume} + \textit{x\_d3}
& 
\makecell{
Conv2D(ks=3, $C_\text{out}$=128, padding=1), ReLU(), \\
Conv2D(ks=1, $C_\text{out}$=128, padding=0), ReLU(), \\
Conv2D(ks=1, $C_\text{out}$=128, padding=0), ReLU(), \\
Conv2D(ks=1, $C_\text{out}$=2, padding=0) \\
}
& \textit{out} & $H/4 \times W/4 \times 2$ \\
\hline
\end{tabular}
\end{center}
\caption{G-Net architecture. Using the cost-volume and the feature-map from D-Net as input, G-Net updates the initial depth probability distribution by estimating $\Delta \mu / \sigma$ and $\sigma^{\text{new}}/\sigma$. Similar to D-Net, we use linear activation for $\Delta \mu / \sigma$, and the modified ELU function \cite{other-ELU}, $f(x) = \text{ELU}(x) + 1$ for $\sigma^{\text{new}}/\sigma$ to ensure positive variance and smooth gradient.}
\label{table:gnet}
\end{table*}

\begin{table*}[t]
\setlength{\tabcolsep}{2.6pt}
\begin{center}
\begin{tabular}{c|c|c|c}
\hline
\textbf{Input} & \textbf{Layer} & \textbf{Output} & \textbf{Output Dimension} \\
\hline
\textit{x\_d3}
& 
\makecell{
Conv2D(ks=3, $C_\text{out}$=128, padding=1), ReLU(), \\
Conv2D(ks=1, $C_\text{out}$=128, padding=0), ReLU(), \\
Conv2D(ks=1, $C_\text{out}$=128, padding=0), ReLU(), \\
Conv2D(ks=1, $C_\text{out}$=4$\times$4$\times$9, padding=0) \\
}
& \textit{out} & $H/4 \times W/4 \times (4\times4\times9)$ \\
\hline
\end{tabular}
\end{center}
\caption{Architecture of the learned upsampling layer. The output, which has the shape of $H/4 \times W/4 \times (4\times4\times9)$ is reshaped into $H \times W \times 9$, and softmax is applied to the last channel. Then, the depth of each pixel in the full resolution is calculated as the weighted sum of the $3\times 3$ grid of its coarse resolution neighbors \cite{other-RAFT}.
}
\label{table:up-net}
\end{table*}

Tab. \ref{table:dnet}, \ref{table:gnet} and \ref{table:up-net} show the architecture of D-Net, G-Net and the learned upsampling layer.

\clearpage

{\small
\bibliographystyle{ieee_fullname}
\bibliography{egbib}
}

\end{document}